\documentclass[preprint,12pt,authoryear]{elsarticle}

\usepackage{amssymb}
\usepackage{hyperref}

\journal{Expert Systems with Applications}

\begin{document}

\begin{frontmatter}



\title{Improving Clustering on Occupational Text Data through Dimensionality Reduction}


\author[1]{Iago Xabier Vázquez García}
\ead{iago.vazquez@itcl.es}
\fntext[fn1]{\textit{1st Author's Google Scholar:} {https://scholar.google.es/citations?user=xIIVOpYAAAAJ}}

\author[2]{Damla Partanaz}
\ead{damla.partanaz@khas.edu.tr}
\fntext[fn1]{\textit{2nd Author's Google Scholar:} {https://scholar.google.com/citations?user=n1bcwzoAAAAJ}}

\author[2]{E. Fatih Yetkin}
\ead{fatih.yetkin@khas.edu.tr}
\fntext[fn1]{\textit{3rd Author's Google Scholar:} {https://scholar.google.com.tr/citations?user=nntoUOQAAAAJ}}


\affiliation[1]{organization={ITCL Technology Center},
            addressline={López Bravo St., 70}, 
            city={Burgos},
            postcode={09001}, 
            state={Castilla y León},
            country={Spain}
            }
\affiliation[2]{organization={Kadir Has University, Management Information Systems Department},
            addressline={Civali, Fatih, Fatih}, 
            city={Istanbul},
            postcode={34083}, 
            state={Istanbul},
            country={Turkey}             
            }

\begin{abstract}
{In this study, we focused on proposing an optimal clustering mechanism for the occupations defined in the well-known US-based occupational database, O*NET. Even though all occupations are defined according to well-conducted surveys in the US, their definitions can vary for different firms and countries. Hence, if one wants to expand the data that is already collected in O*NET for the occupations defined with different tasks, a map between the definitions will be a vital requirement. We proposed a pipeline using several BERT-based techniques with various clustering approaches to obtain such a map. We also examined the effect of dimensionality reduction approaches on several metrics used in measuring performance of clustering algorithms. Finally, we improved our results by using a specialized silhouette approach. This new clustering-based mapping approach with dimensionality reduction may help distinguish the occupations automatically, creating new paths for people wanting to change their careers.}
\end{abstract}




\end{frontmatter}


\section{Introduction}
\label{intro}
Due to the incredible advancement in computer power over the last decades, a new age known as Industry 4.0 has begun. Industry 4.0 transforms how businesses produce, enhance, and distribute their goods. The Industrial Internet of Things (IIoT), cloud computing, data analytics, and machine learning (ML) are among the cutting-edge technologies which are incorporated into various manufacturing processes \citep{IIoT}. Consequently, their impact involves automating tasks, thereby replacing or displacing numerous jobs across various sectors. This replacement, has been around for a while. In the 1950s, the term "computer" described the human workers who performed manual calculations; today, no one is needed to operate the elevators, like it was needed in the past \citep{kasparov2017deep}. Ha Joon defines the contemporary problem as having the pressure of displacement on highly-educated professionals \citep{chang2015economics}. Brynjolfsson also states that ML differs from previous waves of automation since high-wage professions are also affected by it \citep{brynjolfsson2017can}. Frey and Osbourne's study has shown a \%47 risk of displacing labor \citep{frey2017future}. This analysis was done before the pandemic era, when afterward autonomy gained more importance and even raised. Despite this rapid technological development, there is also hope in this problem. According to Brynjolfsson, there are no occupations in which all of its tasks will be replaced by full automation \citep{brynjolfsson2020economics}. When some of those tasks are automatized the remaining tasks can be merged with other occupations'  tasks and become a new occupation. Many examples exist in today's workplaces, e.g., database governance and cloud engineers. In addition, the people whose occupations will be replaced soon might continue with a similar occupation that will not be automatized in the near future. To do that, they would need to gain new skills since in those other occupations, there will be some tasks that their old occupations did not have. Hence, we are developing a recommendation portal that will recommend various online courses to gain new skill sets so that they can switch their occupations with another that will not be automatized in the near future.

 
In this article, we proposed a reliable and accurate clustering mechanism to build a model that will figure out the similarity between occupations required to work in the background of a dynamic survey -which we will explain how and why- designed for obtaining the necessary dataset to perform such a recommendation. The first condition for such a recommendation portal is to a sufficient and reliable dataset. The Occupational Information Network (O*NET) is a commonly used open dataset in this field \citep{article, Deming2017, Xu2021}. There are 1016 occupations listed in the database with their definitions, corresponding tasks, and skills. O*NET was first released and shared with the public in 1998; the Standard Occupational Classification (SOC) taxonomy has been updated and made open to the public several times over the years, with the last version being released in 2019 with 23 major groups \citep{onet2023taxonomy}. Nonetheless, this dataset is insufficient for designing such a recommendation platform since (1) there is no task-skill match in it, (2) no data about the risk of the tasks can be automatized, and (3) it is country specific. Therefore, to enrich this dataset, new data should be collected from various companies and their employees via a well-designed survey.
 
This survey should match the skills required to perform the tasks for each occupation and to collect the possibility of these tasks to be automatized. 
For this reason, it is crucial that the task lists of the people who will participate in the survey are determined correctly and that the questions are presented to them in this way. Since the people participating in the survey will not be asked to write their tasks one by one manually, in order to facilitate this process, it was planned to use the O*NET dataset to structure the survey.
Although there are pre-defined tasks connected with the occupations in O*NET, the task lists cannot be directly gathered from the O*NET database once the survey participant entered their occupations, since there is no universal truth, the definitions of each occupation may vary from company to company, country to country, and hence their tasks may differ. So, suppose the tasks are gathered directly from O*NET, once the survey attendees select their occupations. In that case, the survey will show questions regarding the same tasks for each participant from different firms and countries, even if their tasks are not the same. That can easily lead to such problems as (1) collecting data with wrong labels, (2) collecting incomplete data, and (3) collecting incorrect data. In addition, there might be some occupation titles that O*NET might not have, so those survey attendees can not proceed that way.

\begin{figure}[!h]
    \centering
    \includegraphics[width=320pt]{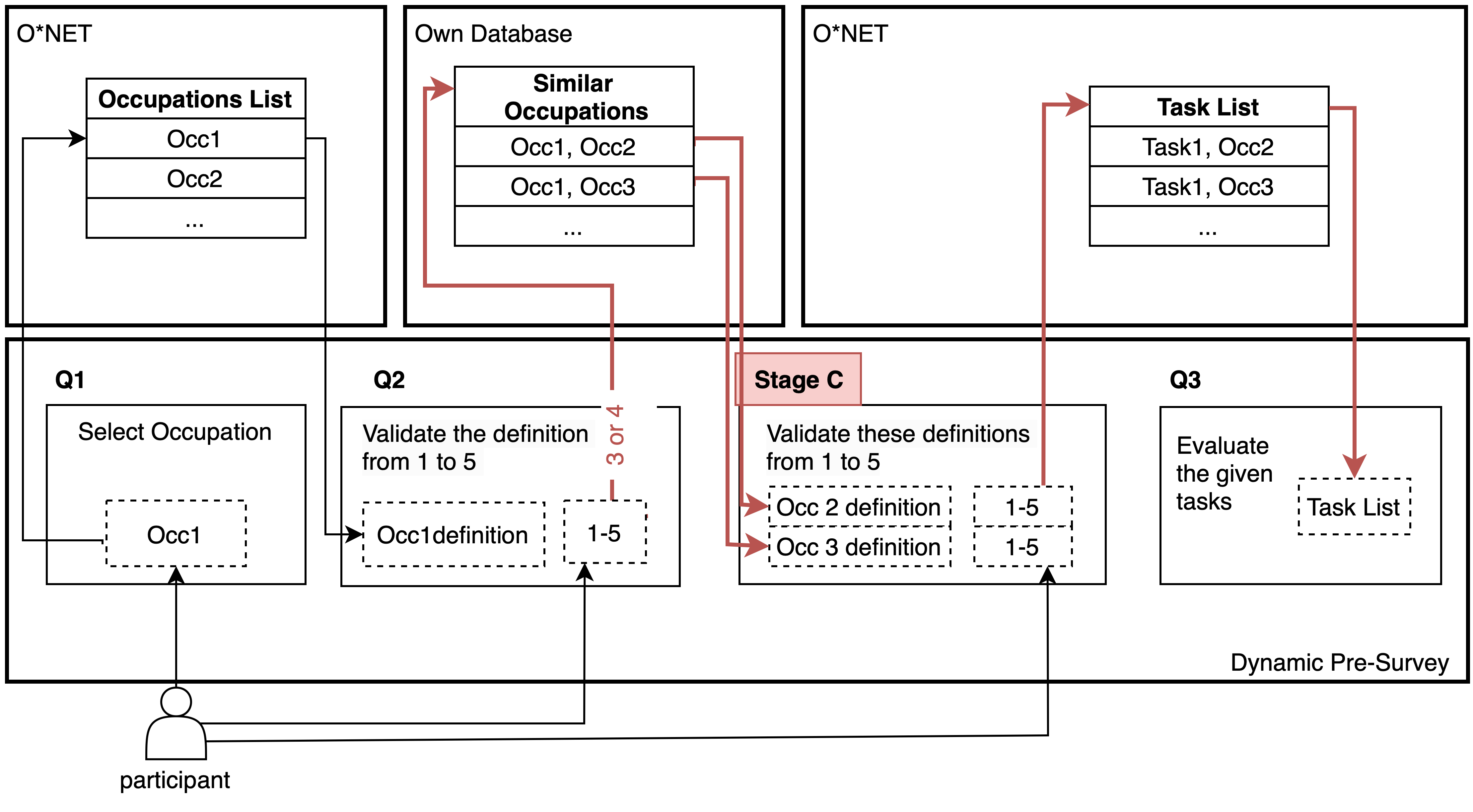}
    \caption{The structure of the dynamic pre-survey: A special flow. Suppose the survey attendee evaluates the given definition for their occupation(Occ1) as 3 or 4, similar occupations of Occ1 are gathered and listed in matrix form with their definitions in Stage C. In that case, survey attendees evaluate all the given similar occupations, and accordingly, in Q3, the tasks are combined and shown to the survey attendees to evaluate them. Finally, all tasks exceeding the threshold(3) will be recorded for that user and linked to the task questions in the questionnaire.}
    \label{fig:dynamic}
\end{figure}

 To solve this problem, a dynamic pre-survey is designed based on the O*NET dataset, which will be conducted before the questionnaire, only to detect the tasks of the survey participants and link them to the task-related questions in the questionnaire. 
This dynamic pre-survey will start by checking whether the definition of the survey attendee's occupation is aligned with the definition of O*NET for the given survey attendee's occupation. For the scenario where the survey attendee's occupation is listed on O*NET already, a special flow can be seen in Figure \ref{fig:dynamic}. Once the survey attendees select their occupations for the first question(Q1), the survey gathers its definition according to the O*NET for the second question(Q2), and survey attendees are asked to validate that definition according to their situation in the firms they are working. If they enter 5, stage A, which matches the definition of the O*NET, the survey gathers the task lists from O*NET for the survey attendees to validate them and finish this pre-survey. If they enter 1 or 2, stage B, means the users' definitions and the O* NET are far away. Hence, the survey attendees should manually select all similar occupations from the drop-down menu, one by one, and evaluate them according to their definitions, again from 1-5. According to their scores, those occupations' tasks will be combined and shown to the survey attendees to evaluate as the last step. In the special flow, where all its steps are shown in Figure \ref{fig:dynamic}, if they enter 3 or 4, it means the survey attendee's and the O*NET's definitions are not so far away from each other, and O*NET's similar occupations can be in line with the similar occupations that the survey attendee would choose manually (like in Stage B). So, in this case, to make the process more user-friendly, the survey lists all the similar occupations of the survey attendees' occupations, with their definitions, for the survey attendees to evaluate again. 

In this article, we will focus on two aims. (1) making this dynamic pre-survey user-friendly by directly showing similar occupations to the survey attendees while they do not have to select them one by one manually. (2) In addition, planning that in such a way that in the future, for the scenarios where the users' occupation is not listed in the O*NET, the survey can continue by automatically clustering the given definition by the survey attendee.

By these aims,
we proposed an accurate and reliable model to cluster occupations by their definitions. To do that, we used O*NET's SOC as the ground truth and trained our model accordingly. The reason for this approach is for the scenarios in which survey participants' occupation will not be in the O*NET, and even though that, by getting the definition from that participant, it will be able to detect that given occupation's cluster and thoroughly gather the most reasonable task list again. However, for now, we only focused on the definitions of the occupations listed in O*NET and used several method combinations to develop an accurate and reliable clustering mechanism.

The rest of the paper is organized as follows. In Section \ref{related}, the related literature is presented. The proposed methodology is explained in detail in Section \ref{metodology}. Section \ref{experiments} presents and discusses the proposed approach's numerical results. Finally, the last section concludes the study and discusses future directions.



\section{Related Works}
\label{related}

\subsection{Semantic Analysis}

Word embeddings were a branch of very popular models several years ago. Models like Global Vectors for Word Representation (GloVe) \citep{pennington2014glove} or Word2Vec \citep{journals/corr/abs-1301-3781} are able to compute similarity between pairs of words. They assign one vector to every word with enough appearance frequency in a text corpus, in such a way that they preserve semantic relationships among the corresponding words. Then, the cosine between these vectors works as a measure of the similarity between the original pairs of words. A cosine value close to 0 indicates that the words are unrelated, and a cosine value close to 1 indicates that they are strongly semantically related. However, word embedding models are unable to make a difference among several different meanings of the same word. This problem was solved by the Embeddings from Language Model (ELMo) \citep{peters-etal-2018-deep}. ELMo is a bidirectional Long Short Term Memory (LSTM) that takes account of the whole sentence in which a word is used to generate the corresponding word embeddings.

Eventually, the development of the transformer architecture \citep{vaswani2017attention} allowed the development of the Bidirectional Encoder Representations from Transformers model (BERT) \citep{devlin-etal-2019-bert}. BERT combines the transformer architecture with the ELMo’s bidirectional structure. On the other hand, its self-supervised training strategy prevents the need for manual labeling of training datasets, so larger amounts of data can be used to train it in an easier way than in the usual supervised strategy. Models fine-tuned from BERT achieved the best performance in several Natural Language Processing (NLP) tasks \citep{Choi2021EvaluationOB}, like question answering or sentence classification. However, this model was not still able to compute the similarity between sentences. To compute similarity metrics between two sentence embeddings, it is required for the two vectors to have the same size. As different sentences have a different number of words, this is not a trivial problem. To solve this gap, Sentence-BERT (S-BERT) \citep{reimers-gurevych-2019-sentence} was developed. This model adds a mean pooling layer to the BERT embeddings, so they keep a prefixed size. Then, it uses a Siamese or Triplet Network architecture to fine-tune the original BERT model on pairs or triplets of sentences to predict if the input sentences are similar, neutral or dissimilar among them. Finally, the embeddings generated by the model at the mean pooling layer’s output can be used to compute the similarity between two sentences. Cosine, Manhattan and negative euclidean distances can be used to perform this calculation with similar results. 
In recent years, Sentence-BERT models were used successfully in several tasks, such as retrying articles to understand the background of a given one \citep{Deshmukh2020IRBERTLB}, enriching a lexicon with new entries \citep{Pontes2022UsingCS}, or grading short answers on not seen questions \citep{Condor2021AutomaticSA}.
\subsection{Dimensionality Reduction} 
Real-life data, including text data, generally have a complex and non-linear structure. Hence, most machine learning models suffer from the well-known issues called empty-space phenomena and the curse of dimensionality \citep{survey}. This concept roughly means that once the number of the features increased, the volume of the space also expanded, and the samples became more sparse in such a large volume of space and prevent the recognition of the patterns in data. Application of dimensionality reduction as a post-processor is not new in language transformers \citep{dimRedLang}. However, most of the studies are focusing on the implementation of linear dimensionality reduction methods such as Principal Component Analysis (PCA) for reducing the storage needs of the large language models and the computational time of the similarity calculations. On the other hand, linear methods, such as PCA, are primarily considering the distribution of the data and do not preserve the geometry of the data points in general \citep{survey}. Therefore, our focus in this study is to measure the effect of well-known Manifold Learning approaches such as Laplacian Eigenmaps (LE) \citep{le}, Locally Linear Embedding (LLE) \citep{lle}, Locality Preserving Projections (LPP) \citep{lpp}, Neighborhood Preserving Embedding (NPE) \citep{npe}, T-distributed Stochastic Neighborhood Embedding (t-SNE) \citep{tsne}. These methods have two main computational bottlenecks that prevent their usage in very large datasets: 1) computation of the neighborhood graph, 2) computation of the required number of the eigenpairs of related eigenproblem. There are some recent studies focusing on to reduce manifold learning algorithms' complexities \citep{megaman}. Although not having a-priori information about the intrinsic dimension of the data can be considered as another important disadvantage of the dimensionality reduction approach, very recent studies can be found in the literature for the efficient estimation of the intrinsic dimension \citep{idEst}. 

\section{Methodology}
\label{metodology}
Workers' occupations are far away from following a standard nomenclature. Despite the existence of some databases that try to establish a standard, such as O*NET and European Skills, Competences, Qualifications and Occupations (ESCO){\footnote{\url{https://esco.ec.europa.eu/en}}}, companies may work with different definitions for similar occupations. Moreover, O*NET and ESCO were designed for the U.S. and the European labour markets, respectively. Their listed occupations are different, and they cannot be mapped to occupations in 
third countries in a trivial way. This is an obstacle when studies about labour market are conducted in different countries and companies. On the other hand, recent developments in NLP allow measuring the similarity between a pair of sentences, by means of S-BERT \citep{reimers-gurevych-2019-sentence} models {and similar techniques}. This technology, applied to the occupations definitions, allows knowing how similar different occupations are. Then, by applying clustering techniques, it is possible to find clusters that represent very similar occupations, in such a way that they can be used as standard occupations.

\subsection{Methods}
\label{methods}
BERT \citep{devlin-etal-2019-bert} is a neural network based on the Transformer architecture. It is able to generate word embeddings, taking into account the context where they were used. This way, BERT is able to make a difference between different meaning of the same word.
By using a mean pooling layer and fine-tuning techniques, BERT models can be applied to calculate similarity between pairs of sentences. Some of the most widely used BERT models in the NLP community were selected for this study\footnote{All mentioned BERT models can be found at \url{https://huggingface.co/}.}:

\begin{enumerate}
\item[A)] \textit{distilbert-base-uncased} \citep{Sanh2019DistilBERTAD}: Lighter version of the original BERT model. It was trained by using a soft-label distillation loss and a cosine embedding loss to mimic the original BERT model.
\item[B)] \textit{nreimers/MiniLM-L6-H384-uncased} \citep{wang2020minilm}: Lighter version of the original BERT model, distilled from it by mimicing the self-attention modules.
\item[C)] \textit{sentence-transformers/all-MiniLM-L6-v2}: S-BERT model generated from the model B. The fine-tuning process was executed on pairs of questions and answers, pairs of sentences with same meaning, and pairs of titles and articles.
\item[D)] \textit{sentence-transformers/multi-qa-distilbert-cos-v1}: S-BERT model fine-tuned on pairs of questions and answers. It was developed from the pretrained model A.
\item[E)] \textit{sentence-transformers/paraphrase-MiniLM-L3-v2} \citep{reimers-2019-sentence-bert}: It is the original S-BERT model. Fine-tuned on pairs of equivalent sentences from the original BERT model.
\item[F)] \textit{bert-base-uncased:} This is the full original BERT model, without any distillation process.
\end{enumerate}
The listed BERT models are applied to the descriptions of the occupations available in the O*NET database. For each description of each occupation $o$, a S-BERT model generates a vector $v(o)$ of a fixed size $m$, via the application of a mean pooling layer. This layer is also applied to the outputs of the non-S-BERT models, in the same way that the S-BERT ones, to force their outputs to have the same size. These final outputs were also normalized to have euclidean norm 1. This step allows us to get similar results when using either cosine or euclidean distance, allowing at the same time the comparison among the performance of clustering on BERT outputs and reduced data if euclidean distance is used. Then, for each pair of occupations $(o_1,o_2)$, the euclidean distance between $v(o_1)$ and $v(o_2)$ is calculated. This way, the matrix $M$ is built in such a way that every element $M_{i,j}$ corresponds to the distance between $v_i$ and $v_j$, where $(i,j)$ is a pair of occupations from O*NET. 

We employed well-known linear and non-linear dimensionality reduction algorithms to observe the effect of the dimension on clustering accuracy and reliability. Although we primarily focused on non-linear approaches such as LE, LLE, LPP, NPE, and t-SNE, we also employed some classical approaches such as Principal Component Analysis (PCA) and Multidimensional Scaling (MDS) for comparative reasons. Note that, we are using all these dimensionality reduction methods without optimizing their computational complexities. The main bottleneck of these approaches is commonly known as the underlying eigenproblem. However, in our experiments, we are using an occupational database, and it has a relatively small number of samples (approx. 1000 occupations). Therefore the eigenvalue problem arising from the experiments is relatively small and feasible. 

We have employed well-known clustering methods to the outputs of reduction methods and BERT models. They were selected to take into account algorithms with different behavior. We apply the classical algorithms k-means \citep{electronics9081295} and k-medoids \citep{ARORA2016507}, which are only able to find clusters with a simple shape; dbScan \citep{Bhattacharjee2020}, which is able to find clusters with more complex shapes than the previous ones, but it cannot deal well if different clusters have different density, and spectral clustering, that is able to deal with complex structure and a variable density.\citep{6321045}. Which clustering method is better can provide insights into the underlying structure of the embeddings generated by BERT models.

\subsection{Evaluation Metrics}
Occupations in the O*NET database are classified hierarchically according to the SOC structure{\footnote{A more detailed information can be found at \url{https://www.onetcenter.org/dl_files/Taxonomy2019_Summary.pdf}}}. Thus, 1016 occupations are classified across three hierarchical levels, from the 
general to the detailed categories. The wider category, named "Major Group", contains 23 different classes. The clusters obtained from our experimentation are evaluated against the major group category in the SOC structure in the following way, similarly to the approach exposed in \citep{Rand1971ObjectiveCF}:
(1) If two occupations are in the same major group and in the same cluster, the pair is considered as a \textit{true positive} (TP), 
(2) if two occupations are in different clusters and belong to different major groups, they are counted as a \textit{true negative} (TN),
(3) if two occupations are in the same cluster but they belong to different major groups, they are counted as a \textit{false positive} (FP), and
(4) if two occupations are in different clusters but they belong to the same major group, they are counted as a \textit{false negative} (FN).


Once TP, TN, FP and FN are known, the performance of the obtained results is evaluated by using the following metrics:


\textit{Accuracy (AC)}: \citep{Rand1971ObjectiveCF}
Also called \textit{Rand Index}. Rate of pairs correctly classified. It is calculated as:
\begin{equation}
AC=\frac{TP+TN}{TN+TP+FN+FP}
\end{equation}

AC give values between 0 and 1. Higher values are considered to indicate a better cluster assignment according the ground truth.

\textit{Adjusted Rand Index (ARI):} \citep{ari}
Accuracy adjusted to account for the expected agreement between random assignments. ARI has values up to 1. An ARI of 0 implies that the cluster assignments are not better than a random choice, and higher values imply a better accordance with the ground truth.

\textit{Youden index (YI):} \citep{hughes2015youden} Value between 0 and 1 that 
maximizes the sum of the true positive and true negative rate. It is calculated as:
\begin{equation}
YI=\frac{TP}{TP+FN}+\frac{TN}{TN+FP}-1
\end{equation}

As sensitivity (SE) is defined as the ratio between True Positives and the total number of true values, and specificity (SP) is the ratio between True Negatives and the total number of negative values, Youden index can be rewritten as:
\begin{equation}
YI=SE+SP-1
\end{equation}

Moreover, we also use the following metrics, that does not account for the accordance in the assignment of pairs of elements, but rather perform a measurement of the information shared by the found clusters and the true labels: 

\textit{Mutual Information (MI):} \citep{mi}
Measurement of the amount of information shared by the cluster assignments and the true labels. It is calculated by applying the following equation:
\begin{equation}
MI = \sum_{V\in L}\sum_{U \in C}P(x\in U \land x \in V) \log\frac{P(x \in U \land x \in V)}{P(x \in U)P(x \in V)}
\end{equation}
Here, $L$ is the set of true classes, while $C$ represents the set of found clusters. $P(x \in D)$ represents the probability for the element $x$ of belonging to the set $D$. Meanwhile, $P(x\in U \land x \in V)$ is the probability for an element $x$ of belonging to the class $V$ and to the cluster $U$ at the same time. MI always takes positive values. Higher values imply a better accordance to the ground truth.

\textit{Adjusted Mutual Information (AMI):} \cite{ami}
Measurement of the amount of information shared by the cluster assignments and the true labels, corrected to take into account the number of assignments due to randomness. AMI takes values up to 1. A value of 0 implies that cluster assignment is not better than a random one.

Finally, we employ the \textit{silhouette} metric, \citep{10.1007/978-3-319-62416-7_21} which measurement of how close a point is to its cluster in relation to the closest different cluster. It is calculated as:
\begin{equation}
S(i)=\frac{b(i)-a(i)}{max\{a(i),b(i)\}}
\end{equation}
Here, $a(i)$ is the mean distance from the point $i$ to the other points in the same cluster, while $b(i)$ is the mean distance from the point $i$ to the points in the closest, different from its own one, cluster to $i$. 

silhouette is the only non-supervised metric employed in this article.



Note that, because the number of major groups is higher than two in the SOC structure, the number of pairs such that the two elements belong to the same cluster and the same major group is very small if compared with the total number of pairs. This could make accuracy a misleading metric, as clustering algorithms that generate a lot of clusters with very few elements each would trend to find a higher accuracy. For this reason, we decided to use the Youden index beside the accuracy, that takes into account both the rate of true positives and the rate of true negatives at the same time.


\subsection{Proposed Mechanism}

\begin{figure}[!ht]
  \centering
  \includegraphics[width=300pt]{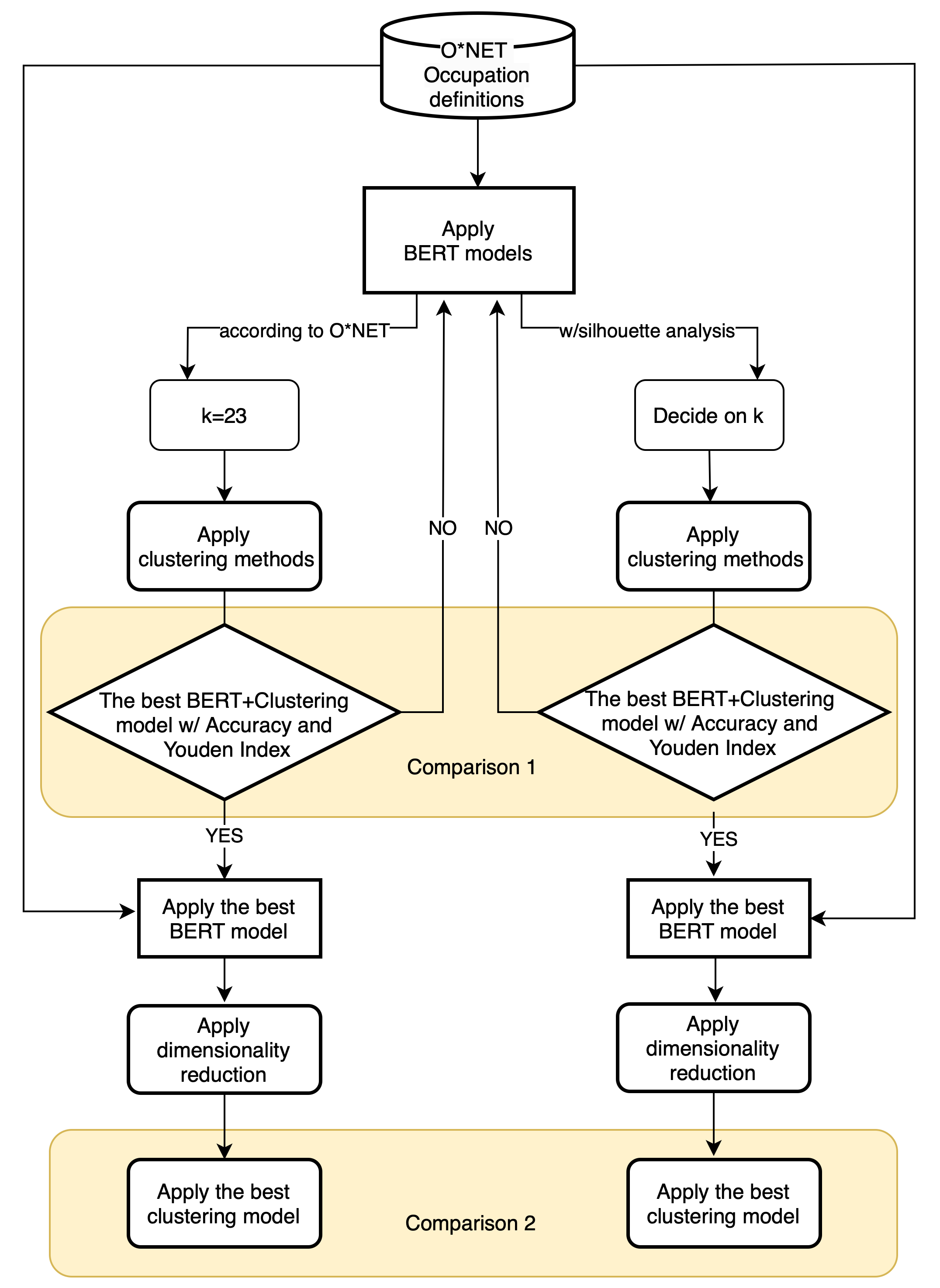}
  \caption{Flowchart of the proposed mechanism.}
  \label{fig:flow}
\end{figure}

For measuring the effect of different post-processing schemes, two experimental pipelines that illustrated in Figure \ref{fig:flow}, are proposed in this paper. The first experimental pipeline does not use the silhouette analysis. In this process, the occupation descriptions in the O*NET database were transformed into vector embeddings by using the five BERT models listed in \ref{methods}, then, the clustering algorithms k-means, dbScan, k-medoids and spectral clustering are used to group the occupations. Although we used both cosine and euclidean distances during the experiments, we depict only the results for the euclidean distance, as the results were roughly similar for both. For clustering algorithms, except for dbScan, the number of clusters to find is set to 23, as this is the number of the different major groups in the SOC Structure. For dbScan, instead, different values of epsilon have been tested, with differences of 0.01 between successive values, and the one that gives a number of clusters closer to 23 is selected.
Finally, AC, ARI, YI, MI and AMI are calculated for each combination of distance, BERT model and clustering algorithm. The best combinations according to each of the listed evaluation metrics are selected for the next step.
At the step 2, the dimensionality reduction methods mentioned in \ref{methods} are applied to the two selected models. The mentioned clustering algorithms are applied to the outputs of the dimensionality reduction methods, and AC, ARI, YI, MI and AMI are calculated again to select the best models according these metrics.


In the second experimental pipeline, silhouette analysis is used when clustering is performed. Several different number of clusters to find are tried, and silhouette is calculated for every occupation. The number of clusters that reach the highest mean silhouette value is selected as the optimal to train the clustering model. For dbScan, several epsilon values are tried, instead of setting explicitly the number of clusters to find. Every experiment was run 5 times to reduce the variance in the final performance.





\section{Numerical Experiments}
\label{experiments}
\subsection{Effect of dimensionality reduction in plain form}
\label{plain}
\begin{figure}[!h]
  \centering
  \includegraphics[width=300pt]{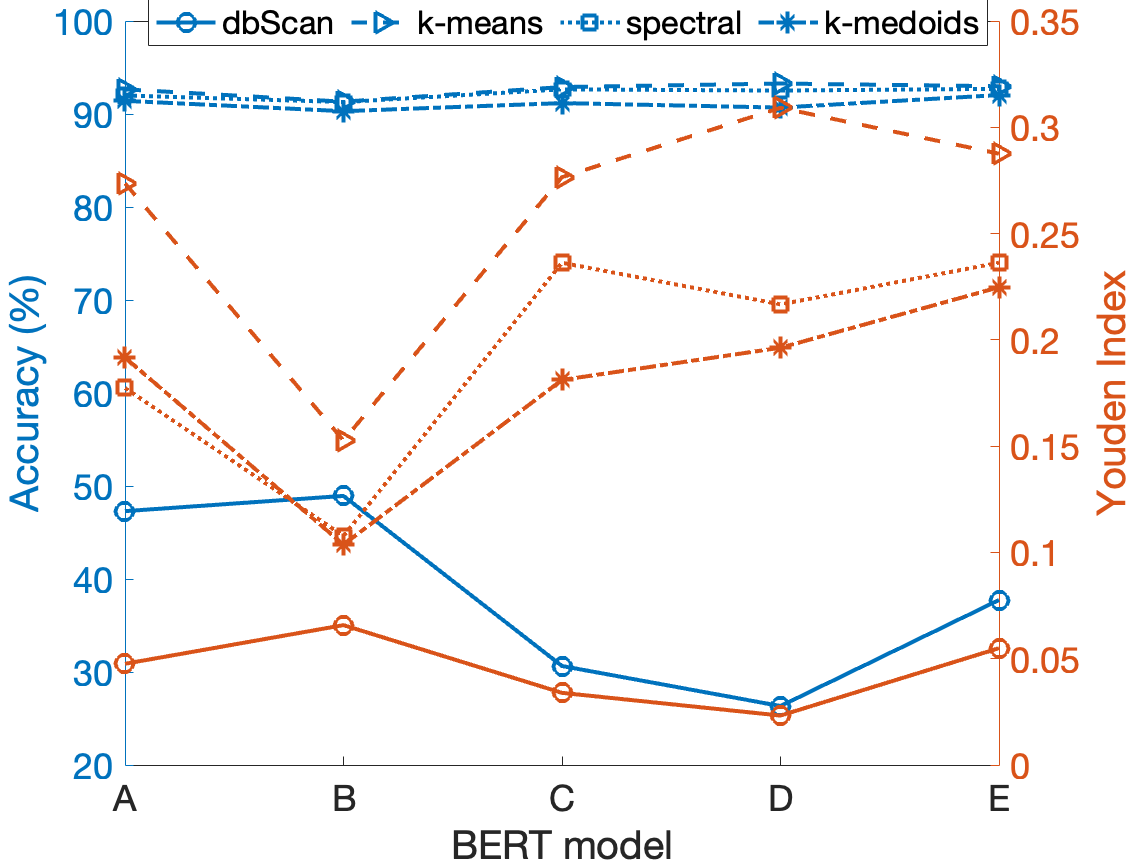}
  \caption{Accuracy and Youden Index plot for various BERT models and clustering approaches without silhouette improvement.}
  \label{fig:nosil_acc_you}
\end{figure}

 As shown in Figure \ref{fig:flow}, the first step of the experimental pipeline is realized to produce distance matrices by using the mentioned BERT models and applying several clustering approaches to find the highest AC, ARI, YI, MI and AMI values at the same time. It is noteworthy that the highest score was obtained by the same model for three of the employed metrics: AC, AMI and MI selected the model \textit{sentence-transformers/multi-qa-distilbert-cos-v1}  (model D) with k-means. Instead, the other two metrics have selected the model \textit{sentence-transformers/paraphrase-MiniLM-L3-v2 } (model E) with k-means. Depicted in Figure \ref{fig:nosil_acc_you}, the accuracy and Youden index obtained for the different BERT models and clustering approaches can be seen. Note that, here we are using $k=23$ as the number of the clusters determined as the ground truth in O*NET database.
 
\begin{figure}[!h]
  \centering
  \includegraphics[width=300pt]{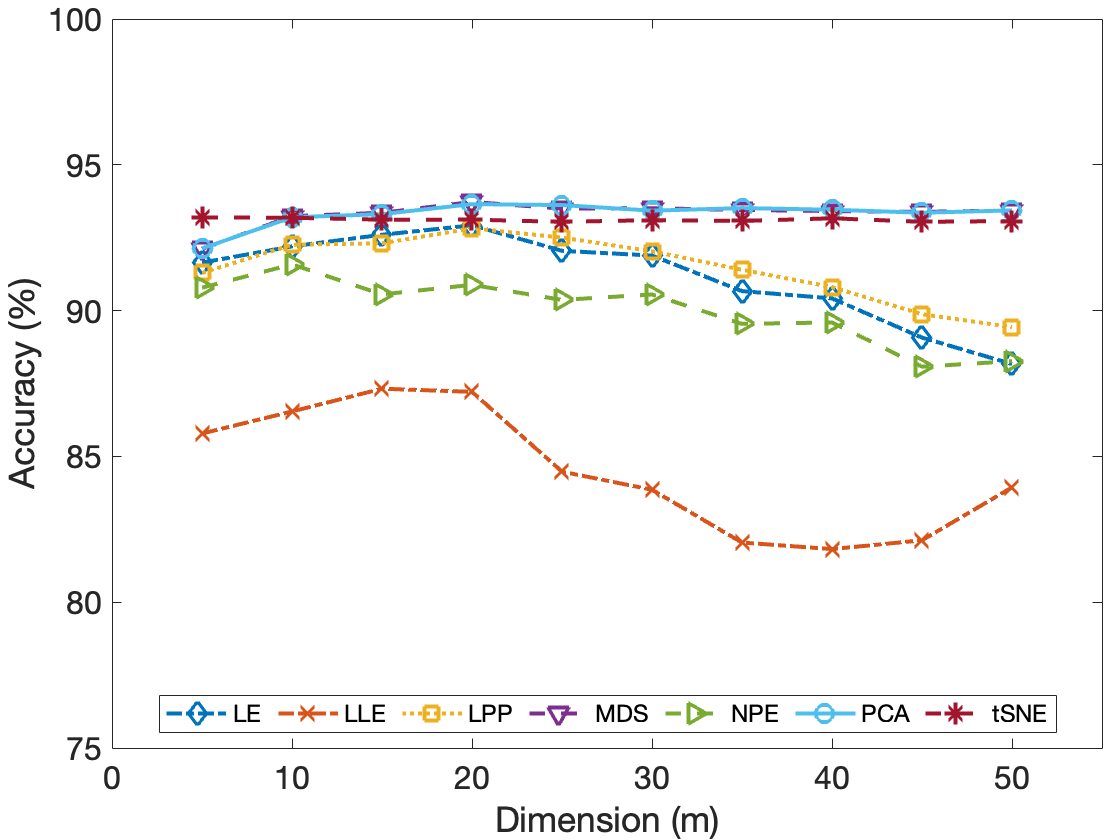}
  \caption{Accuracy behaviour of BERT model D and k-means clustering approach with different dimensionality reduction techniques.}
  \label{noSil_acc}
\end{figure}

\begin{figure}[!h]
  \centering
  \includegraphics[width=300pt]{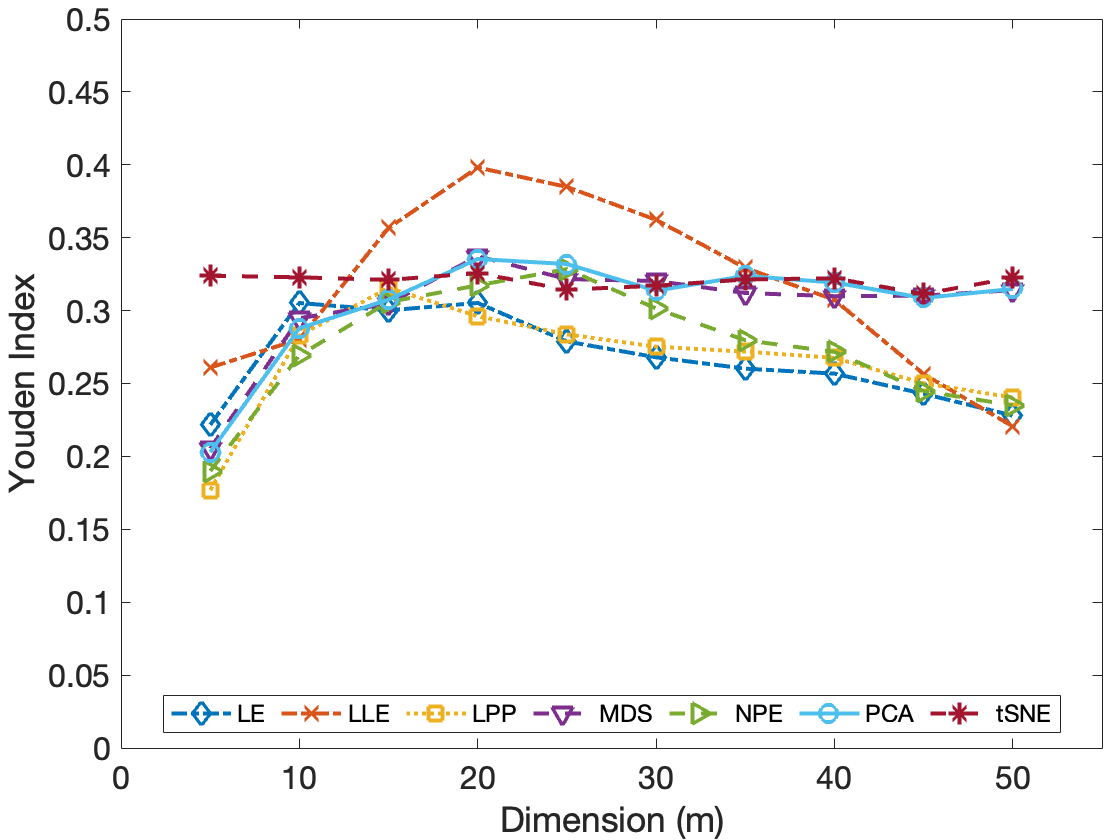}
  \caption{Youden index behaviour of BERT model D and k-means clustering approach with different dimensionality reduction techniques.}
  \label{noSil_youden}
\end{figure}

 To observe the effect of the dimensionality reduction techniques, we re-run the same clustering approaches with the reduced data matrices into different dimensions ($m$). Note that in the original data matrix, model D has $m=768$, while model E has $m=384$. We have applied reduction techniques to reduce their dimension in a range between 5 to 50, 5 by 5, and for dimensions 100, 200, 300, and, additionally, for model E, 400, 500, 600 and 700.  Best models found after applying dimensionality reduction to D+k-means and E+k-means, according to each metric employed, are summarised in the Table  \ref{table2}.  Moreover, the accuracy and Youden index values obtained for model D+means are depicted in the Figures \ref{noSil_acc} and \ref{noSil_youden}.

In the Table \ref{table2}, for each model selected from step 1, the best reduction method and reduced dimension, according to each metric, are listed. Columns $m_1$ and $m_2$ display the original dimension of vector embeddings and the reduced dimension, respectively, while columns $\sigma_1$ and $\sigma_2$ represent the metrics for the performance of the original and reduced embeddings, according to the metric used to select them, specificied in the \textit{Metric} column. Ultimately, for each of these metrics, the superior model found is recognized by the highest $\sigma_2$ value reported in this table.

 The clear observation from the Figures \ref{noSil_acc} and \ref{noSil_youden} is the stable behavior of the t-SNE method. Even though it is not the best choice according to any of the five metrics used, as it can be seen in the Table \ref{table2}, its accuracy behavior seems does not depend on the dimension of the data. On the other hand, LLE method especially for $m=20$ has a great improvement effect on the Youden index. By contrast, it is significantly reducing the accuracy of the clustering for almost all $m$ values. It is noteworthy that, for both selected models, the improvement achieved by the use of reduction methods is low for every metrics except for the Youden index.
 
Another important observation is that dbScan has the worst results for the selected BERT models across the whole performed experimentation. dbScan is able to find complex structures in general, but it fails when different clusters have different density. Possibly, the distribution of embeddings generated by BERT models have a variable density. However, it is also notorious that k-means and k-medoids, which only can deal with simple structures, but are less sensitive to different densities than dbScan, have provided better results. It is also worth to mention the the poor performance of the BERT model B if compared with the other BERT models. Besides, the model C, fine-tuned from the model B, has a similar performance to the obtained by the other BERT models. Maybe, the distillation through mimicing the self-attention modules is good for fine-tuning, but does not keep the similarity between sentences as well as the BERT model A does.
 

\subsection{Effect of dimensionality reduction with silhouette improvement}

The second path in the proposed experiment pipeline includes the application of silhouette analysis for selecting the best number of the clusters for each BERT method and clustering algorithm combination. 

Our observation is that the silhouette analysis improves all the metrics for almost all combinations except the dbScan. The highest improvement is produced for the Youden index. Using this metric to select the best models leads us to models with a better balance between sensitivity and specificity. On the other hand, AC and MI select models with high specificity and low sensitivity, while ARI and AMI retrieve models with higher sensitivity and less specificity than RI and MI, but with a worse balance between these two metrics than Youden index.

As the number of major groups is high, there is an imbalance between the number of pairs of occupations belonging to the same major group and the number of pairs of occupations belonging to different major groups. As most pairs of occupations belong to different major groups, finding a high number of clusters leads to fewer pairs whose occupations belong to the same cluster, which can help to increase specificity at the expense of sensitivity. In fact, the optimal number of clusters found by AC, RI, AMI and ARI is similar to the number of major groups, while the number of clusters found when Youden index is employed is significantly lower. It is noteworthy that AC and MI get the highest specificity values, and the lowest sensitivities, while ARI and AMI, their versions corrected to erase the effect of randomness in matching clusters and major groups, get higher sensitivities and lower specificities.



\begin{figure}[!h]
  \centering
  \includegraphics[width=300pt]{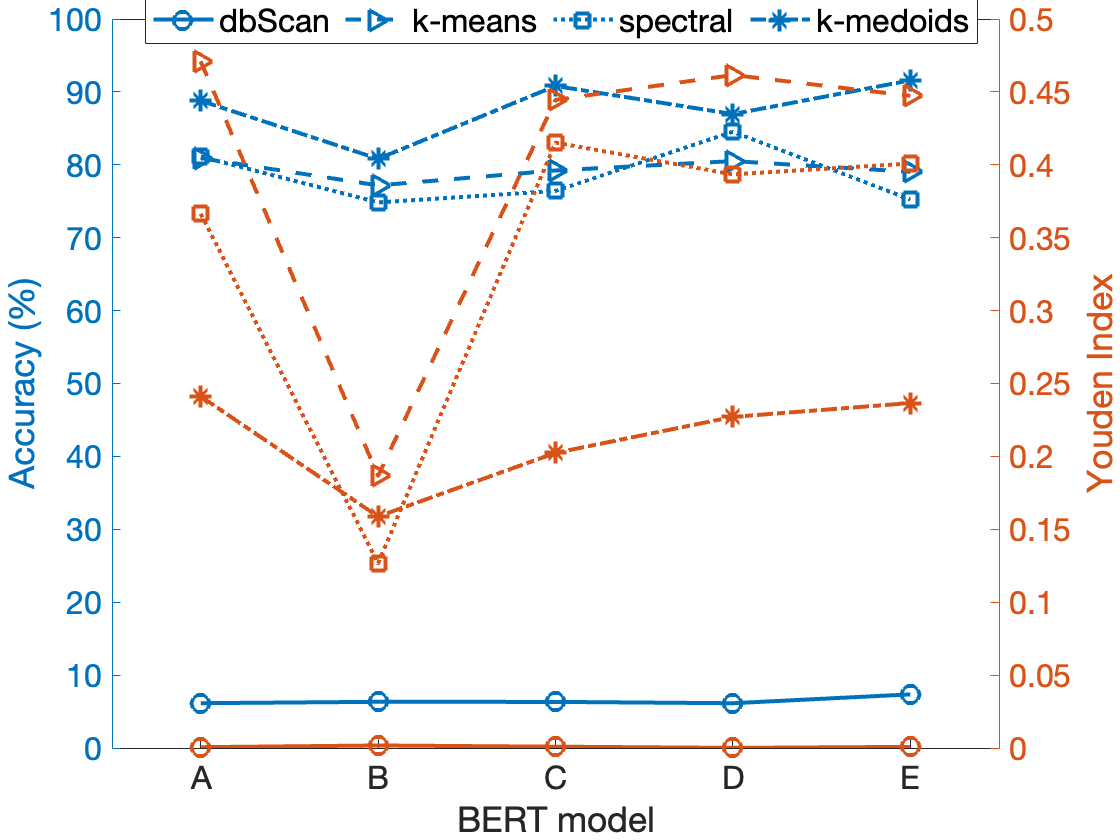}
  \caption{Accuracy plot for various BERT models and clustering approaches with silhouette improvement.}
  \label{fig:sil_acc_youden}
\end{figure}
After the implementation of silhouette analysis to determine the best number of the clusters, in contrast to the previous experiments with a fixed value of the number of clusters, we have three different combinations with respect to the employed metrics. Therefore, we repeated the dimensionality reduction experiments for the three cases. 

\begin{figure}[!h]
  \centering
  \includegraphics[width=300pt]{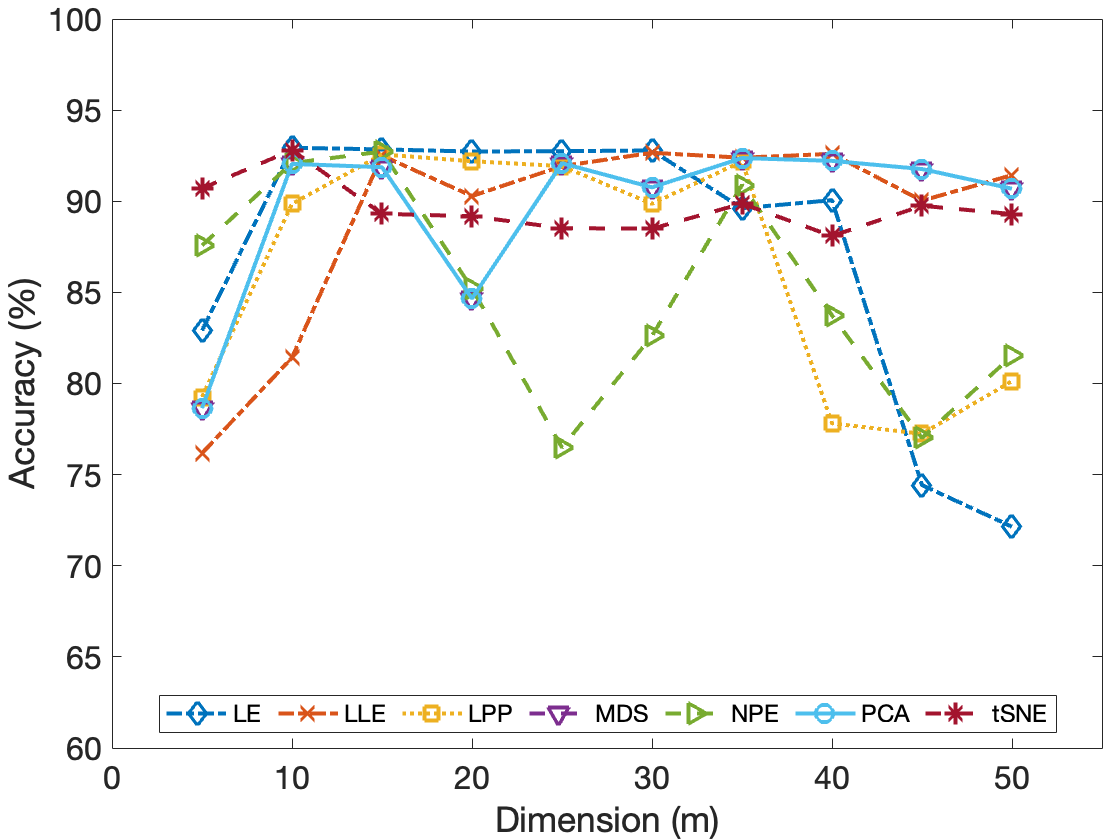}
  \caption{Accuracy behavior of BERT model A and k-means clustering approach with different dimensionality reduction techniques.}
  \label{acc_sil2}
\end{figure}

\begin{figure}[htbp]
  \centering
  \includegraphics[width=300pt]{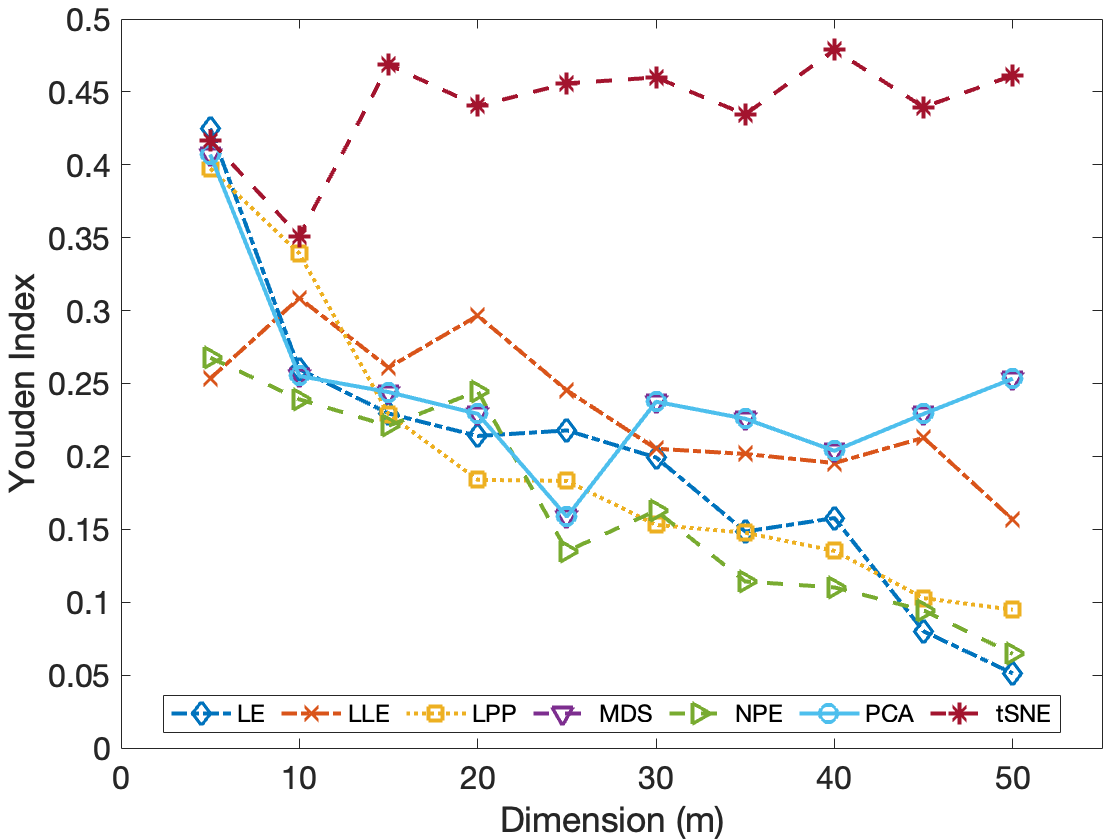}
  \caption{Youden index behavior of BERT model A and k-means clustering approach with different dimensionality reduction techniques.}
  \label{youden_sil2}
\end{figure}

The best model for the highest Youden index is the distilbert-base-uncased (model A) with k-means clustering. For AMI, it is the model D with k-medoids, and for the rest of the metrics, it is E+k-medoids. The accuracy and the Youden index plots are shown in Figures \ref{acc_sil2} and \ref{youden_sil2} respectively, while the best models obtained with this approach are listed in the Table \ref{table1}. This table follows the same structure as Table \ref{table2}, which has been already exposed in Subsection \ref{plain}. 

The first observation that can be made is that to employing silhouette analysis significantly improves the performance of the reduction methods across all metrics. Note that, after the reduction, we again implemented the silhouette analysis to understand the effect of reduction on the number of the optimal number of clusters as well. The best model found according to the accuracy, obtained this time with the BERT model D and the clustering method k-means, has a mean of $k=27.4$ clusters. However, the best model according Youden index has a mean number of clusters of $k=10$, and the best one according to MI, $k=29$, and we obtained $k=16.4$ for the best models found according AMI and ARI. As it can be seen, for the best models, the number of clusters found by the silhouette analysis can vary, but it is roughly similar to the number of classes in the ground truth.



.



\begin{figure}[!h]
  \centering
  \includegraphics[width=300pt]{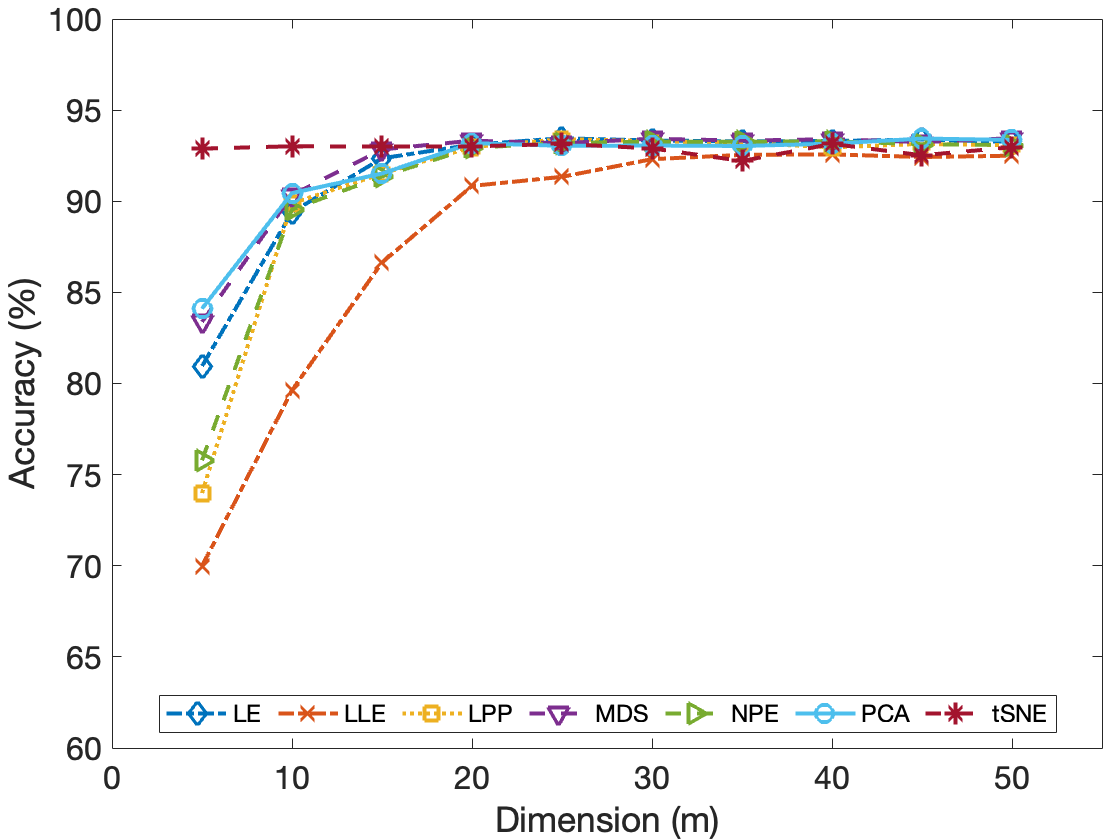}
  \caption{Accuracy behaviour of BERT model E and k-medoids clustering approach with different dimensionality reduction techniques.}
  \label{acc_sil}
\end{figure}

\begin{figure}[!h]
  \centering
  \includegraphics[width=300pt]{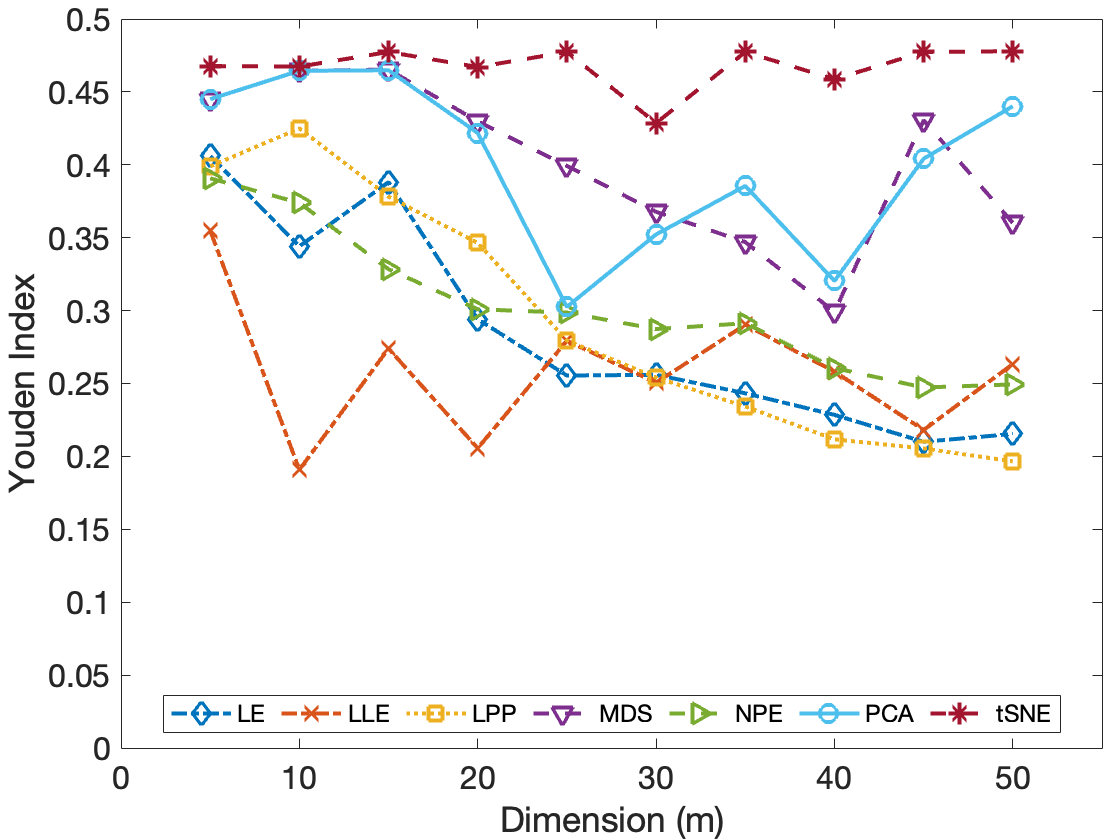}
  \caption{Youden index behaviour of BERT model E and k-medoids clustering approach with different dimensionality reduction techniques.}
  \label{youden_sil}
\end{figure}
\begin{table*}[!h]
\centering
\caption{Improvement with dimensionality reduction with selected methods, withouth silhouette analysis.}
\label{table2}
\begin{tabular}{|ccc|cc|cc|}
\hline
Method & Metric & Reduction & $m_1$ & $m_2$ &  $\sigma_1$ & $\sigma_2$ \\
\hline
D+k-means               &ac&MDS&768&30&0.931&0.936\\ 
D+k-means               &ami&MDS&768&45&0.473&0.498\\ 
D+k-means               &mi&MDS&768&45&1.569&1.64\\ 
D+k-means               &ari&MDS&768&30&0.311&0.353\\ 
D+k-means               &youden&LLE&768&20&0.281&0.391\\ 
E+k-means             &ac&MDS&384&20&0.931&0.935\\ 
E+k-means             &ami&MDS&384&20&0.468&0.481\\ 
E+k-means             &mi&MDS&384&20&1.552&1.592\\ 
E+k-means             &ari&MDS&384&20&0.314&0.348\\ 
E+k-means             &youden&LLE&384&20&0.286&0.351\\ 
\hline
\end{tabular}
\end{table*}
\begin{table*}[!h]
\centering
\caption{Improvement with dimensionality reduction with selected methods, employing silhouette to determine the number of clusters.}
\label{table1}
\begin{tabular}{|ccc|cc|cc|}
\hline
Method & Metric & Reduction & $m_1$  & $m_2$  &  $\sigma_1$ & $\sigma_2$ \\
\hline
E+k-medoids               &ac&LE&384&10&0.916&0.929\\ 
E+k-medoids               &ami&TSNE&384&10&0.312&0.463\\ 
E+k-medoids               &mi&LE&384&15&1.06&1.55\\ 
E+k-medoids               &ari&TSNE&384&10&0.227&0.343\\ 
E+k-medoids               &youden&TSNE&384&300&0.225&0.475\\ 
D+k-means               &ac&LE&768&20&0.805&0.933\\ 
D+k-means               &ami&TSNE&768&500&0.381&0.493\\ 
D+k-means               &mi&LPP&768&30&0.912&1.674\\ 
D+k-means               &ari&TSNE&768&500&0.206&0.356\\ 
D+k-means               &youden&MDS&768&600&0.459&0.464\\ 
A+k-means               &ac&LE&768&25&0.808&0.929\\ 
A+k-means               &ami&LPP&768&20&0.372&0.469\\ 
A+k-means               &mi&LE&768&25&0.892&1.581\\ 
A+k-means               &ari&LPP&768&20&0.21&0.309\\ 
A+k-means               &youden&TSNE&768&30&0.461&0.471\\ 
\hline
\end{tabular}
\end{table*}

On the other hand, it is interesting to note that the most robust approach is again the t-SNE for the dimensionality reduction. This time, t-SNE has been the reduced method that achieves the highest performance with the highest frequency. t-SNE is a good candidate to use as a post-processor for the BERT model since it improves every metric significantly. Although t-SNE is one of the most expensive manifold learning approaches, as we discussed earlier, the dataset size is not very large for the occupational definitions dataset. 

Moreover, it is always possible to use some sophisticated techniques such as MegaMAN \citep{megaman} to reduce the execution times for large scale datasets. 
 

\section{Conclusion and Future Work}

In this study, we analyze the approaches for improving the accuracy and reliability of the clustering algorithms on occupational data. The study's overall objective is to use the proposed mechanism as a tool for a broader study focusing on mapping the occupational definitions from different countries and firms. Consequently, this clustering mechanism will be used for creating a dynamic pre-survey and to structure and obtain the necessary and sufficient occupational dataset to design a platform that makes some recommendations to people who need to change their occupations because of automation. We employed silhouette analysis to improve the accuracy and reliability metrics (for selecting the optimal number of clusters) and the linear/non-linear dimensionality reduction approaches in a pipeline structure. The experimental study shows that applying these approaches significantly impacts the clustering results' reliability. The approach will be embedded in a survey structure to make the data collection scheme more flexible and reliable. Also, the consistent behaviour of the t-SNE reduction method makes it a promising technique to apply to S-BERT embeddings to solve clustering problems on text data. 
Moreover, in the experiments, we used a ground truth defined by the experts in the O*NET taxonomy structure. In future studies, this approach will also be implemented to the more extensive and unlabeled data available in O*NET, such as tasks and skills, aiming for it to work with the unseen definitions of the new occupations that are not listed in the O*NET.
On the other hand, the results obtained make us think that the application of methods similar to the ones described in this article could substitute, in the future, multiple choice answers in surveys, allowing the development of a new field in social science.


\section*{Declarations}

\begin{itemize}
\item {\bf{Funding:}} This work was supported by the European Union (EU) Research and Innovation Staff Exchange programme through the Reshaping labour force participation with Artificial Intelligence (AI4LABOUR) Project under Grant  101007961. 
The work of Damla Partanaz and E. Fatih Yetkin was supported by Scientific and Technological Research Council of Turkey (TÜBITAK)-1001 under Project 120E281. Other than that, the authors have
no relevant financial, ethical, or non-financial interests to disclose.
\item {\bf {Availability of data and code:}} : We obtained the data from a publicly available repository, namely from O*NET: \url{https://www.onetonline.org/}. The code is
developed as a part of an ongoing project but can be shared upon request.

\item {\bf{Authors' contributions:}} All authors contributed to the study conception,
methodology, and design. Algorithmic design, data curation and visualization were performed by Iago Xabier Vázquez García, Damla Partanaz, and E. Fatih Yetkin. All authors read
and approved the final manuscript, and consent for publication

\item{\bf{Conflict of interest:}} On behalf of all authors, the corresponding author states that there is no conflict of interest.

\end{itemize}



\bibliographystyle{apalike} 

\bibliography{main}





\end{document}